\documentclass{article}
\usepackage[nonatbib, preprint]{neurips_2020}
\usepackage{mathtools}
\usepackage{amsmath, amssymb}
\usepackage{tabularx}
\usepackage[utf8]{inputenc} 
\usepackage[T1]{fontenc}    
\usepackage{hyperref}       
\usepackage{url}            
\usepackage{booktabs}       
\usepackage{amsfonts}       
\usepackage{nicefrac}       
\usepackage{microtype}      
\usepackage[backend=biber, sorting=none]{biblatex}
\usepackage[pdftex]{graphicx}
\usepackage{float}
\usepackage{subcaption}
\usepackage{booktabs}
\usepackage{color}
\usepackage{comment}

\title{Selective Eye-gaze Augmentation To Enhance Imitation Learning In Atari Games}

\author{
  Chaitanya Thammineni\\
  \texttt{sthammin@buffalo.edu} \\
  \And
  Hemanth Manjunatha \\
  \texttt{hemanthm@buffalo.edu} \\
  \And
  Ehsan T. Esfahani \\
  \texttt{ehsanesf@buffalo.edu}\\
  Human In the Loop System Laboratory \\
  University at Buffalo, Buffalo, NY, 14260 USA\\
}

\begin{document}

\maketitle

\begin{abstract}
This paper presents the selective use of eye-gaze information in learning human actions in Atari games. Vast evidence suggests that our eye movement convey a wealth of information about the direction of our attention and mental states and encode the information necessary to complete a task. 

Based on this evidence, we hypothesize that selective use of eye-gaze, as a clue for attention direction, will enhance the learning from demonstration.
For this purpose, we propose a selective eye-gaze augmentation (SEA) network that learns when to use the eye-gaze information. The proposed network architecture consists of three sub-networks: gaze prediction, gating, and action prediction network. Using the prior 4 game frames, a gaze map is predicted by the gaze prediction network which is used for augmenting the input frame. The gating network will determine whether the predicted gaze map should be used in learning and is fed to the final network to predict the action at the current frame. 

To validate this approach, we use publicly available Atari Human Eye-Tracking And Demonstration (Atari-HEAD) dataset consists of 20 Atari games with 28 million human demonstrations and 328 million eye-gazes (over game frames) collected from four subjects. We demonstrate the efficacy of selective eye-gaze augmentation in comparison with state of the art Attention Guided Imitation Learning (AGIL), Behavior Cloning (BC). 

The results indicate that the selective augmentation approach (the SEA network) performs significantly better than the AGIL and BC. Moreover, to demonstrate the significance of selective use of gaze through the gating network, we compare our approach with the random selection of the gaze. Even in this case, the SEA network performs significantly better validating the advantage of selectively using the gaze in demonstration learning.
\end{abstract}

\section{Introduction}
The most common form of human augmentation (guidance) in the learning frameworks is to learn \emph{policy} directly from human actions. In comparison to \textit{reinforcement learning}, the \textit{imitation learning (IL)} framework has shown significant advantages as they do not required handcrafted reward functions \cite{judah2014imitation}. By learning directly from human actions, IL can reduce the huge cost of learning from scratch \cite{silver2016mastering, vinyals2019grandmaster}. Moreover, \textit{human in the loop learning} in IL where human attention can be used to reduce the size of the state and/or action space to guide the IL. For instance, in visual learning tasks, the gaze position indicates a human's immediate attention to process urgent state information. Several researches have successfully utilized the eye gaze-maps to guide the learning process \cite{zhang2019atari, nikulin2019free, saran2020efficiently, zhang2018agil}. In these works, the predicted gaze heat-map is used to select the critical features in a given state resulting in higher accuracy in imitating human actions \cite{li2018eye, zhang2018agil}. Also, incorporating the human attention model into behavioral cloning has shown to improve the Atari game's performance by 115\% \cite{zhang2018agil}. Nonetheless, the effective use of eye-gaze data remains unexplored. In this work, we investigate the selective use of eye-gaze information to enhance the behavior cloning in the Atari platform.

\begin{figure}[b]
    \centering
    \includegraphics[width=0.87\linewidth]{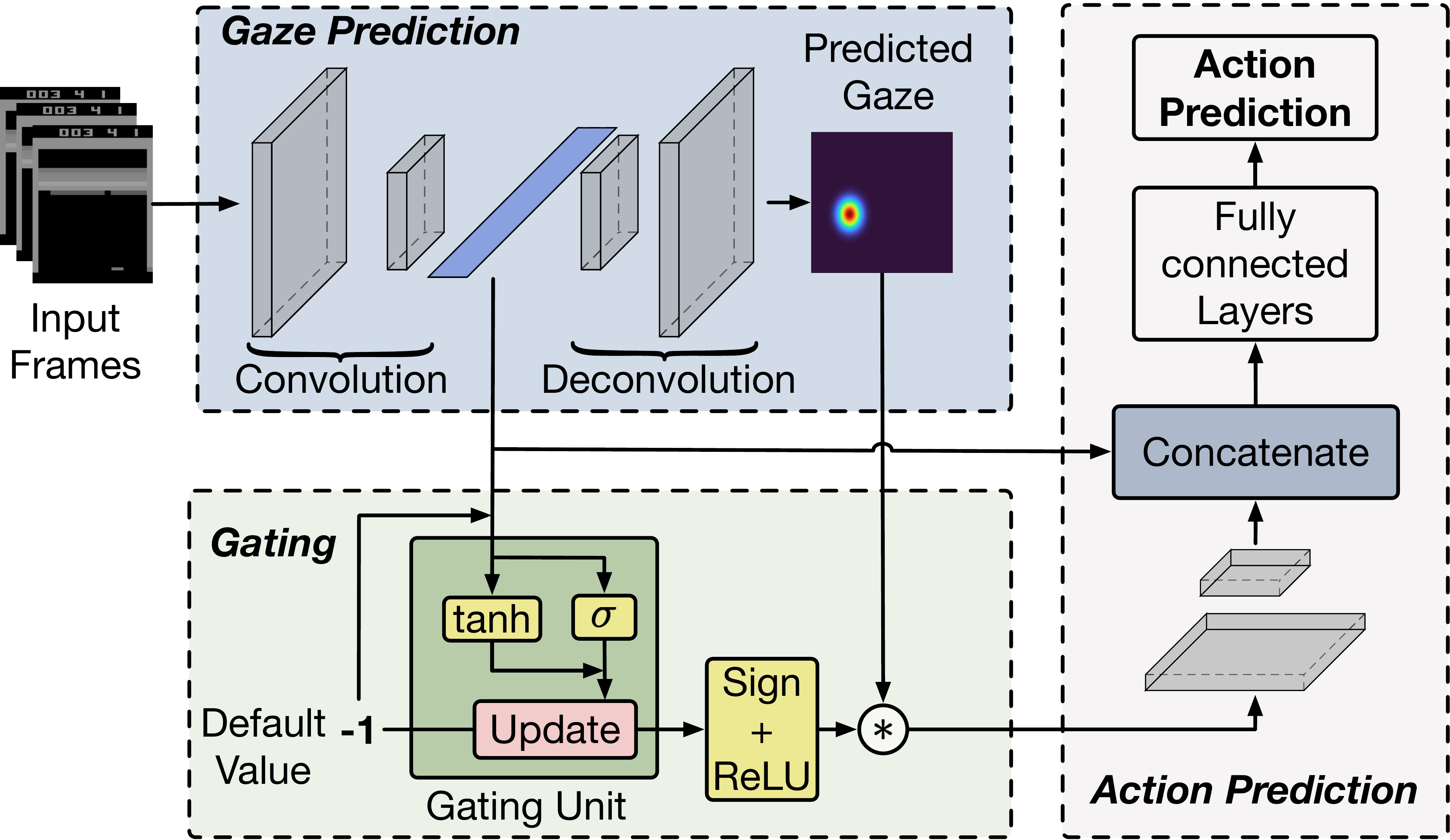}
    \caption{Architecture of selective eye-gaze augmentation (SEA) network. The network has three modules: a) Gaze prediction network, which predicts eye gaze given four frames of the game. b) Gating network that learns when to use the eye-gaze. c) Action prediction network which learns the action mapping from the game frame embeddings and eye gaze information.}
    \label{fig:architecture}
\end{figure}

The rationale for use of eye-gaze is based on decades of evidence that strongly suggests that action selection is facilitated by attention \cite{neumann1987beyond, tipper1997selective, castiello1999mechanisms, cisek2007cortical, petrosino2013selective}. However, not all eye-movements provide information on attention and action. For instance, during eye-saccades no visual information is gathered and the perceptual attention can occur with eye-saccades and similarly fixation on an object does not guarantee that attention is directed \cite{zhao2012eye}. Consequently, over the course of a task, the eye-movements should be selectively processed to understand the instances which are relevant to perform an action. To this end, our main idea is to augment the input frame with the human-gaze only when required and, in any other cases, use an unaugmented input frame. 

For example, an organism usually selects a subset of information by directing their sensory organs towards specific stimuli (over attention) and internally focusing on the particular part of these specific stimuli (covert attention) \cite{petrosino2013selective} to act upon or to select an action among the available set of actions \cite{gibson2014ecological, miller1992electrophysiological, petrosino2013selective}.

Such behavior integrates deeply with our daily activities, such as reaching tasks \cite{land1999roles}, sports, and driving \cite{ahlstrom2011processing}. Moreover, many studies have shown that the eye movements (providing attention) and motor demonstrations are an inter-weaved phenomenon where the visual system extracts the necessary information to complete a task \cite{land1999roles, gredeback2015eye, flanagan2003action}. Arguing in the same lines, we hypothesize that selective augmentation of eye-gaze (thus selective attention) information should provide vital information about the action itself and should enhance the performance of the action imitation system. For this purpose, we propose a neural network architecture that learns when to use the eye-gaze information selectively.

The network architecture (Figure \ref{fig:architecture}) for augmenting the eye-gaze data has three sub-networks: gaze prediction network, gating network, and action prediction network. The \textit{gaze prediction network} takes four frames of the game as an input and predicts the eye-gaze distribution (gaze map) over the last frame in the sequence. This predicted gaze map is used for augmenting the input frame. The \textit{gating network} is used to specify whether the predicted gaze map is used in learning or not. We can achieve this by multiplying the binary output (0 or 1) from the gating network with the predicted gaze map. The \textit{action prediction network} uses information from two channels to predict the action at the current frame. The two channels are embedded input frame information and gated gaze-map. If the gate output is 0, essentially only the input frame information is used for action prediction. There are two loss functions corresponding to gaze prediction and action prediction, and the gaze-prediction loss is \textit{independent} of action prediction loss. Thus, we can decouple the training process and train the gaze-prediction network separately. Therefore we can use a pre-trained gaze-prediction network while training the action prediction network. 

Our main contribution is a  selective eye-gaze augmentation (SEA) network that automatically learns when to use the eye-gaze information for better action prediction. We demonstrate the efficacy of SEA on publicly available Atari-HEAD dataset, which consists of eye-movements of the subjects during the gameplay. The proposed framework is shown to outperform the state of the art Attention Guided Imitation Learning (AGIL) on the same dataset which uses in using eye-gaze information to learn the human actions in Atari. 

\section{Related Work}
Recently, the use of eye-gaze in guiding imitation learning is gaining traction. For example, \cite{barrett2018sequence} showed that by utilizing the human attention from eye-tracking as an inductive bias in recurrent neural network (RNN), the performance could be dramatically increased. The study showed that RNN regularised by human attention improved sentiment analysis, grammatical error detection, and detection of abusive language. Penkov et al. \cite{penkov2017physical} used eye-tracking to learn a mapping between abstract plan symbols and their physical instances. The study showed that the eye-gaze guided system successfully learns the grounding of abstract plan symbols. Eye-gaze have also been successfully used in driving \cite{palazzi2018predicting}. For instance, Yuying Chen et al. \cite{chen2019gaze} presented a gaze modulated drop-out method in deep driving networks for application in driving. The study showed that the gaze modulated drop method reduced the steering prediction error by 23.5\%. On similar lines in navigation, Yuying Chen et al. \cite{chen2020robot} used the graph convolutional networks with attention learned from the human gaze to navigate a robot through a crowd successfully. The study showed that the eye-gaze guided model performed significantly better than the state-of-art methods. 

The eye-gaze augmentation has also found success in-game platforms like Atari. Zhang et al. \cite{zhang2019atari} introduced a large-scale dataset of human actions in Atari video games with simultaneously recorded eye movements. The study showed that using a learned human gaze model to inform imitation learning resulted in a 115\% increase in in-game performance. The above research-works provide a wide range of applications demonstrating the efficacy of augmenting human gaze information into imitation learning. On similar lines, Akanksha Saran et al. \cite{saran2020efficiently} used gaze cues from human demonstrators to enhance the performance of state-of-the-art inverse reinforcement learning and behavior cloning algorithms, without adding any additional learnable parameters to those models. They showed that augmenting existing convolutional architecture with gaze information, guided the learning agent towards better reward function and policy. Ruohan Zhang et al. \cite{zhang2018agil, zhang2019atari} proposed the Attention Guided Imitation Learning (AGIL) framework, in which a learning agent first learns a visual attention model from human gaze data, then learns how to perform the visuomotor task from human decisions. The framework demonstrated the effectiveness of end-to-end learning of visuomotor tasks guided by attention.

\section{Selective Eye-gaze Augmentation Network}
In this section, we briefly present a description of the dataset used for study and, subsequently, provide details on the architecture of the three sub-networks of the selective eye-gaze augmentation (SEA) network: gaze prediction network, gating network, and action prediction network.

\subsection{Dataset Description}
To study the efficacy of selective eye-gaze augmentation, we have used a large-scale Atari-HEAD dataset \cite{zhang2019atari}, which is collected from four subjects playing 20 different Atari games with varying difficulty levels and game dynamics. During the gameplay subject's eye-movements are recorded using EyeLink 1000 eye tracker at 1000 Hz. The game screen was \(64.6 \times 40.0cm\) (\(1280\times840\) in pixels), and the average distance between the subject and the screen was 78.7cm. The subjects were novices who were familiar with the game environment. The dataset contains 117 hours of gameplay, around 28 million human actions, and 328 million eye-gazes. More information on the game statistics and eye-gaze information can be found in \cite{zhang2019atari}. 

\subsection{Gaze Prediction Network}
\label{subsec:gaze-prediction-network}

The gaze prediction network is adopted from Zhang et al. \cite{zhang2019atari} with some modifications. The input for the network is a game frame of channel \textit{c}, width \textit{w} and height \textit{h}. Since we are using monochromatic images, the channel size is \textit{c} = 1. For the prediction of the gaze over the frame \(i\) at a time instance \(t_i\), we use a history of four frames, i.e., \(i \dots i-3\).

The frames are stacked along the channel dimension to form the input tensor \(X^i \in \mathbb{R}^{c\times w\times h}\). A 2-layer convolution block is used to generate the embedding \(X^i_{em} = ReLU(BN(\boldsymbol{W_c} * X^i))\) where \textit{BN} denotes batch normalization.
This embedding is used in the \textit{gating network} as well as in the \textit{action prediction network}. The embeddings are further deconvoluted (\(\boldsymbol{W_d}\), using a 2-layers deconvolution) to generate the gaze prediction map \(\boldsymbol{E_i} = ReLU(BN(\boldsymbol{W_d} * X^i_{em}))\) over the \(i^{th}\) game frame. We have used softmax with Kullback-Leibler(KL) divergence loss function between the \(\boldsymbol{E_i}\) and true human gaze to train the gaze prediction network. We choose KL divergence because we treated the human gaze over images as a probability distribution (a single Gaussian model); thus, KL was an appropriate measure.

Note that the gaze prediction network parameters are not affected by the action error, and consequently, one can decouple the training of the gaze and action prediction networks. In our implementation, the gaze prediction network is trained first, and then the trained network is used during action prediction training. Another advantage of decoupling gaze and action network is that any pretrained gaze prediction network can be used with/without fine-tuning. Also, this reduces the number of parameters to be trained. The main difference between the network given in \cite{zhang2019atari} and ours is, the gaze input into the action network consists of the top 10 percentile values of the predicted gaze map instead of a Gaussian distribution centered around the gaze point. Detailed explanation is provided in Section \ref{sec:experimental-setup}.

\subsection{Gating Network}
\label{sec:gating-network}
\begin{figure}[t]
        \centering
        \includegraphics[width=0.8\linewidth]{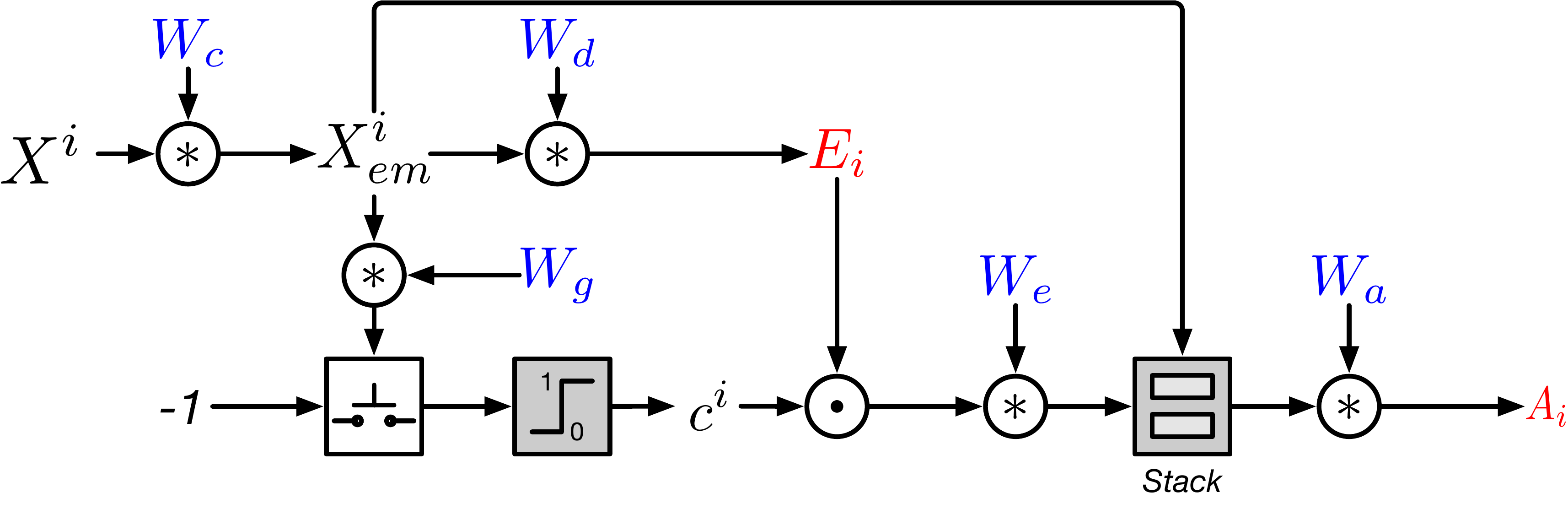}
        \caption{Learnable weights of SEA network. \(\boldsymbol{W_c}\) and \(\boldsymbol{W_d}\) are the convolution and deconvolution weight of the gaze prediction network. The convolution operation produces an embedding (\(X^i_{em}\)) of the input frames. Thus created embedding (\(X^i_{em}\)) is used in gaze prediction, gating network, and action prediction network. \(\boldsymbol{W_g}\) are the weights of the gating network. Lastly, \(\boldsymbol{W_e}\) and \(\boldsymbol{W_a}\) are weights of the action (\textit{A}) prediction network. Note: \(\circledast\) denotes convolution operation and \(\odot\) denotes element-wise multiplication. A fully connected layer is a special case of convolution where the dimension of the kernel size is equal to the input tensor.}
        \label{fig:weights}
\end{figure}

The gating network will identify the instances at which the gaze information should augment the game frames to predict the human actions.
It takes the embeddings (\(X^i_{em}\)) as the input and outputs a binary value of 1 or 0, indicating the use or discarding the gaze information in the action prediction network, respectively. 
The gating function can be modeled as \(c^i=g(\boldsymbol{W_g}*X^i_{em})\) with \(\boldsymbol{W_g}\) as the learnable gating parameter. The gating function \(g\) can be implemented by modifying the popular GRU or LSTM \cite{chung2014empirical} units to be non-recurrent.  
To implement such a behavior, we modify the \(GRU\) unit, as shown in equation \ref{eq:static_gru}. This removes the temporal dependency aspects of the GRU and only preserve the gating functionality.
\begin{equation}
\label{eq:static_gru}
    g(\boldsymbol{W_g}*X^i_{em}) \coloneqq \bigg\{
    \begin{array}{l}
        h^i = GRU(W_g, X^i_{em}, -1)\\
        c^i = ReLU(sgn(h^i))
    \end{array}
\end{equation}
In this formulation, the \(GRU\) unit is implemented with a default hidden state of -1 (\(h=-1\)). The output of the gating network $c^i$ depends on the sign of \(GRU\) unit according to \(c^i=ReLU(sgn(\mathbf{h}))\). The output is then element-wise multiplied with the predicted eye-gaze map \(E_i\), which is subsequently used to augment the input game-frame. Note that the default value of the \(GRU\) (\(h=-1\)), hence the gaze information by default is not utilized for augmenting the game frame unless the \(GRU\) values change.
As shown in Figure \ref{fig:weights}, the \(GRU\) unit weights are influenced by the error in gaze usage and error in action prediction. Consequently, the efficient use of predicted gaze depends on how well the gaze itself is estimated. The dependence of gating network performance on predicted gaze is the desired behavior because a well-estimated gaze can filter not only irrelevant game features, but also enhance necessary features for action prediction.

\subsection{Action Prediction Network}
\label{sec:act-pred-net}
The action network uses two types of embedding to predict the actions. The first embedding is from gating network and calculated as a convolution operation given by \(ReLU(BN(\boldsymbol{W_e} * \boldsymbol{E_i}\odot c^i))\) where the \(c^i\) is the gating network output, and \(\boldsymbol{E_i}\) is the predicted eye-gaze over the input \(X^i\). The second embedding is from gaze prediction network \(X^i_{em}\). The two embeddings are concatenated to form a feature vector to learn the action mapping (Figure \ref{fig:weights}). The concatenated feature vector is subsequently forwarded through a sequence of fully connected layers with learnable weights \(\boldsymbol{W_a}\).
The output of the fully connected layers is one of the feasible actions defined in the Atari game. We use softmax with cross-entropy as the training criterion for action training.


\section{Experiments and Results}
\label{sec:experimental-setup}
This section provides the experimental setup and results of 3 sub-networks of the SEA across six games in Atari-HEAD dataset. Each game in the dataset consists of 20 trials (5 trials per subject) out of which 15 trials were used for training, and five trials were used for inference purposes. The training period was 30 hours over different games. The hyperparameters and training details for three sub-networks of SEA are presented in their respective subsections.

\subsection{Gaze Prediction Network Results}

\begin{figure}[b]
    \centering
    \includegraphics[width=0.86\linewidth]{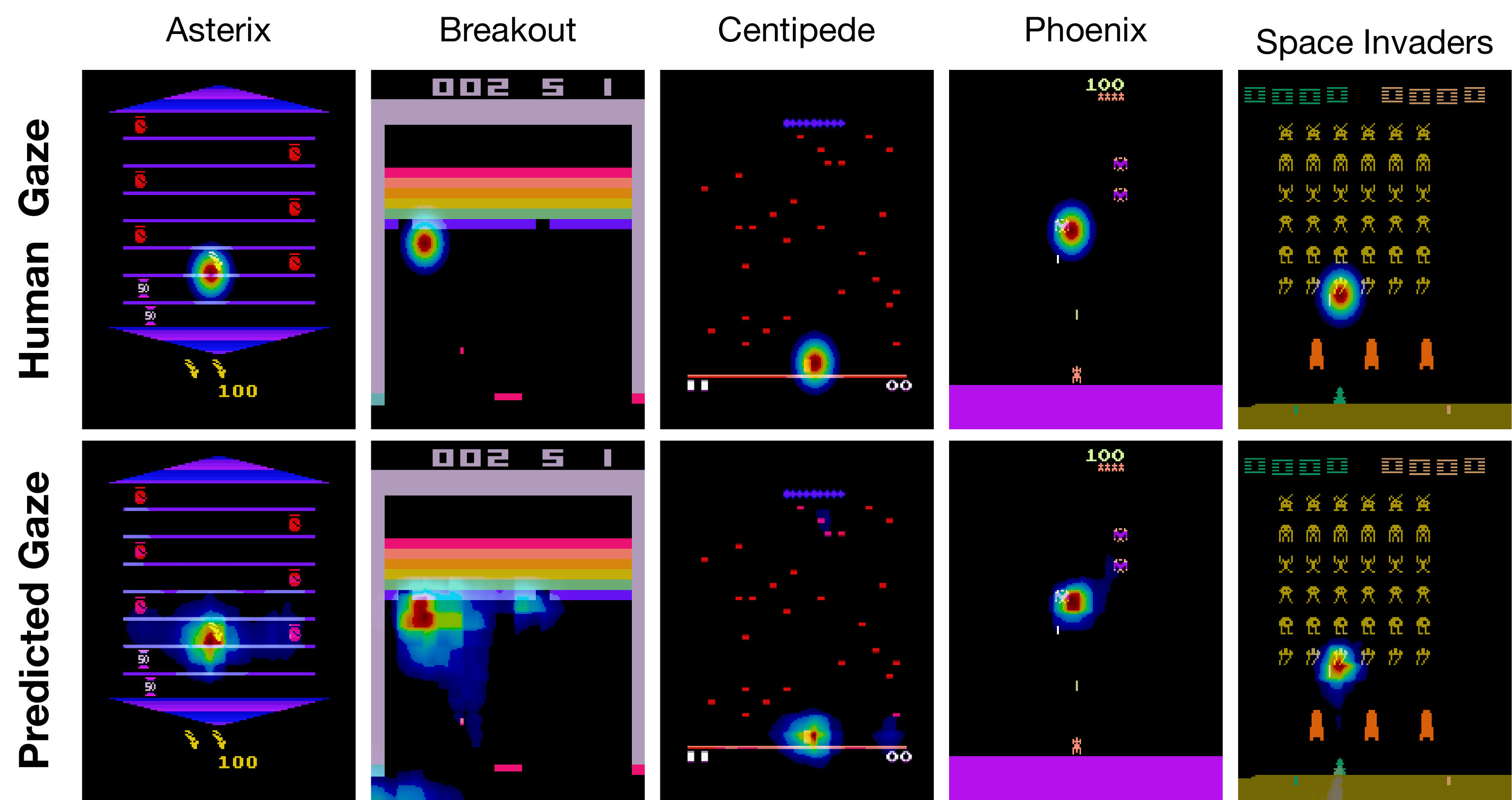}
    \caption{Comparison of human gaze and predicted gaze from gaze prediction network.}
    \label{fig:gazes}
\end{figure}

For gaze prediction, we use a stack of the current frame and the previous three frames. The frames are converted to grayscale and downsampled to \(84 \times 84\) pixel size before stacking. We closely follow the gaze model architecture in \cite{zhang2019atari}, where the gaze is a probability distribution over the 2D image. The probability distribution is calculated by Gaussian estimate with mean and covariance. 
In terms of human gaze samples, we use the last point of the true eye-gazes on the current frame as the ground truth. The true point gaze is converted to a continuous Gaussian probability distribution with a mean-centered at the gaze point and standard deviation \(\sigma\) of one visual degree \cite{meurbaccino2012}. We use KL divergence as the error criterion. Such training results in a point estimate with learned mean and variance. 
Figure \ref{fig:gazes} shows the predicted and actual human gaze in five different games. The spread of the predicted gaze can be seen to be more pronounced in some of the games like Breakout.

\begin{figure}[t]
    \centering
    \includegraphics[width=0.86\linewidth]{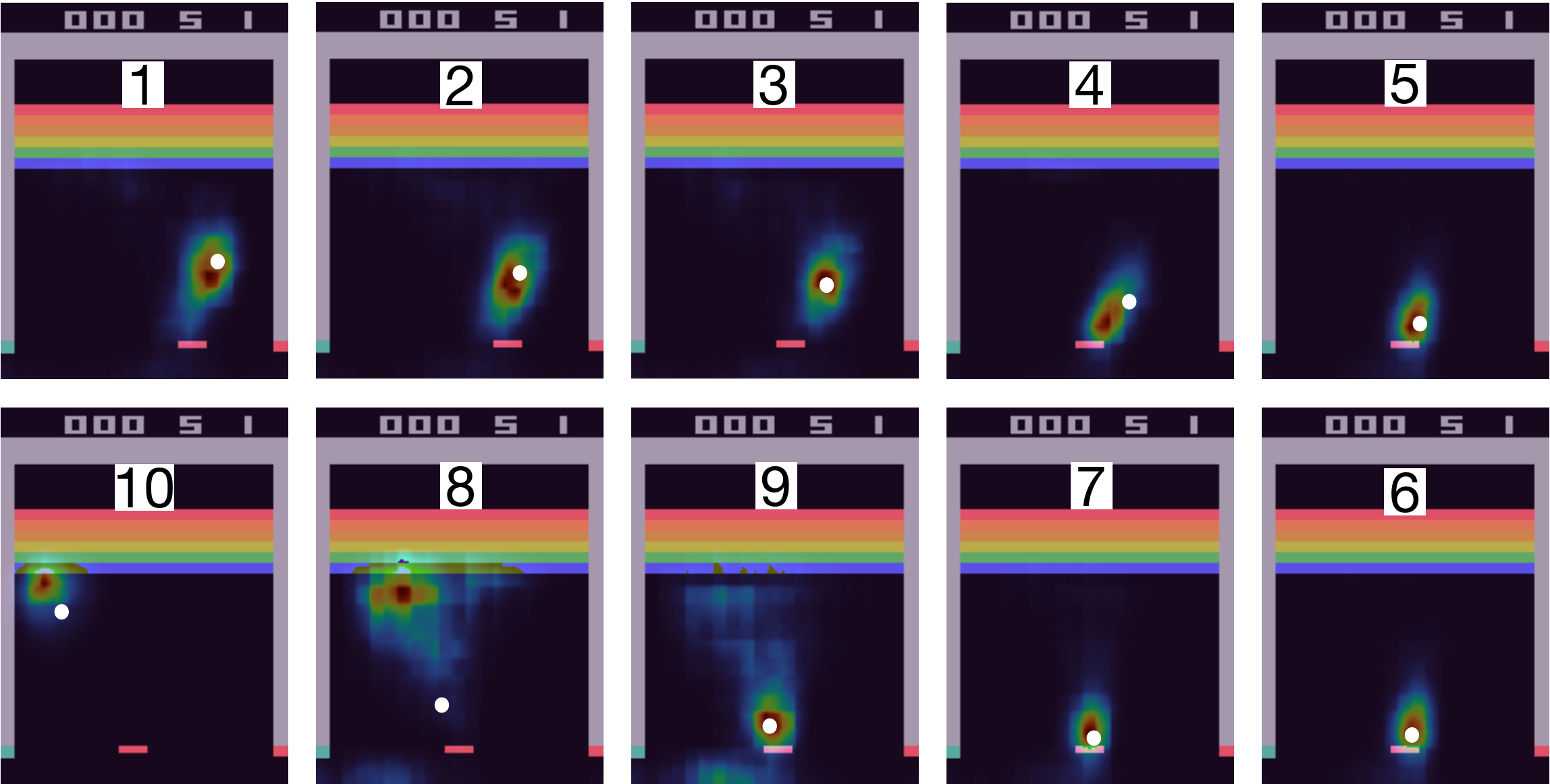}
    \caption{Predicted gaze over a series of frames in the game of Breakout. The predicted gaze shift towards the bricks as soon as the ball leaves the paddle (Ball is highlighted for more clarity).}
    \label{fig:gaze_dynamics}
\end{figure}

Further, we analyze whether the gaze prediction network is capable of understanding the game dynamics. In this regard, Figure \ref{fig:gaze_dynamics} shows the predicted gaze and the ball movement in a Breakout game. The predicted gaze follows the ball closely before hitting the paddle (\ref{fig:gaze_dynamics}, frame 1 to frame 6). As the ball leaves the paddle, the gaze prediction shifts towards the bricks even before the ball reaches the brick. We believe that this behavior is not due to the gaze prediction network architecture, but due to the human gaze data used for training. As the human gaze encodes causal \cite{adams2016dynamic,gerstenberg2017eye} relationship, the gaze prediction network, trained in the human gaze, can learn the causal link to an extent. Such a causal relationship might be hard to learn if we do not use any human data.

\subsection{Gating Network Results}
One of the core tenets underlying the selective gaze utilization is that, by augmenting the game frames with gaze only when required, we can achieve higher performance than using gaze all the time. To achieve this, we employ the gating network that learns to apply the gaze selectively when desired. 

\begin{figure}[h]
    \centering
    \includegraphics[width=\linewidth]{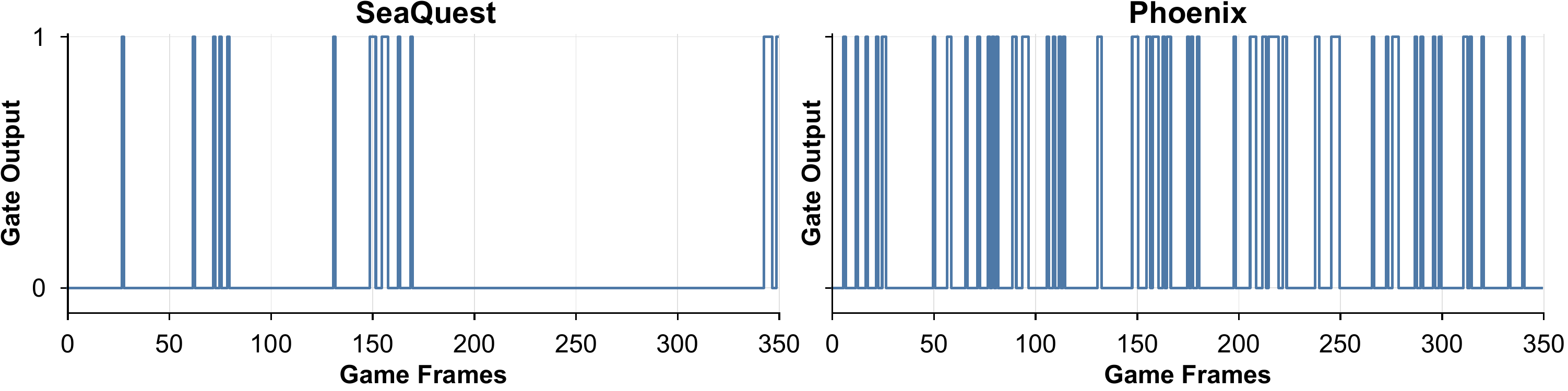}
    \caption{Gate output for SeaQuest and Phoenix at different games frames.}
    \label{fig:gate-games}
\end{figure}

Figure \ref{fig:gate-games} shows the use of gaze over a small duration of gameplay for two games. The gate output of 1 results in gaze being used, and gaze output of 0 results in not being used. We can see that throughout the game, the gating network selects gaze as required, and not all the time. For instance, the gating network output was sparse, and switching happened less frequently for SeaQuest. However, for Phoenix, the switching of gate output is much more frequent.

\begin{figure}[t]
    \centering
    \includegraphics[width=\linewidth]{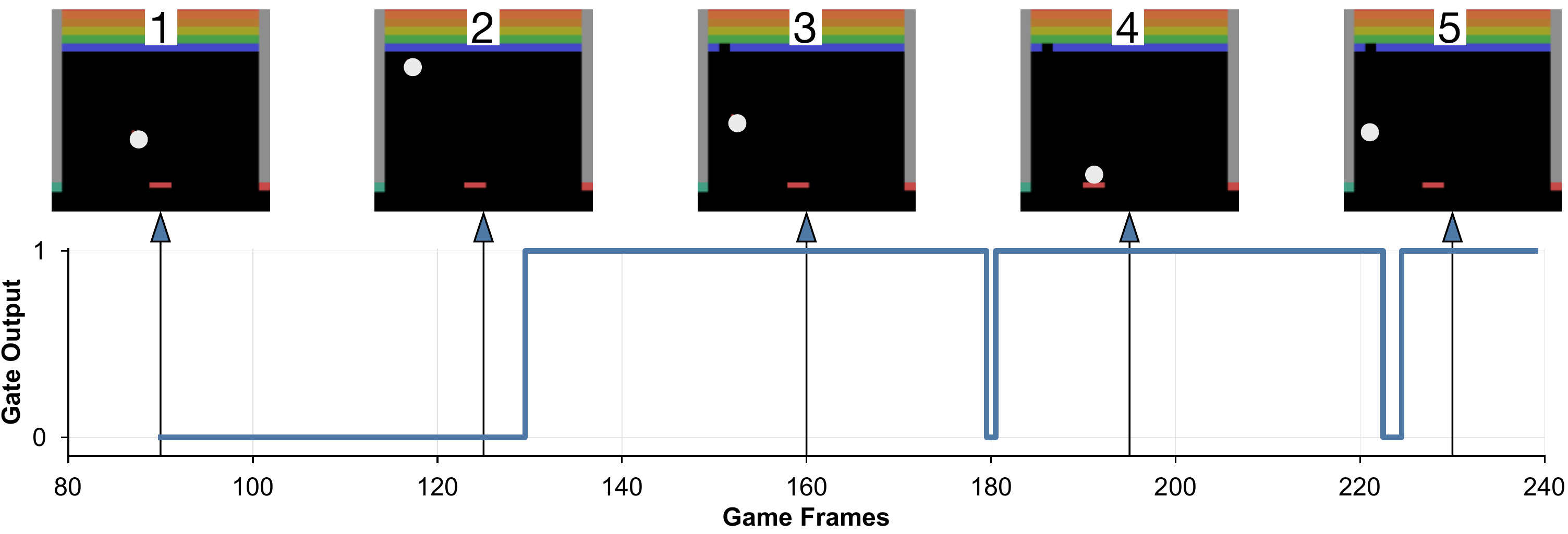}
    \caption{Gate output from gating network with frames at different instances of time. The gate output is 1 when the ball moves towards the paddle. When the ball starts moving towards the paddle, the SEA network starts using the eye-gaze. The gate output is one till the ball leaves the paddle. Note: The results are shown from the part of the game (frame 80 to frame 240)}
    \label{fig:gate-frame}
\end{figure}

To further highlight the dynamics of the gating network, let's consider the results of the Breakout game (Figure \ref{fig:gate-frame}). As it can be seen, the gate output is 0 (thus no eye gaze is used) when the ball is moving away from the paddle (frame 1 in Figure \ref{fig:gate-frame}), at this stage no action is needed. As the ball starts moving towards the paddle (frames 2 and 3) after hitting the brick, the paddle needs to adjust by moving right or left. At this moment, the gate output turns on using the gaze data. The gate remains open until the ball hits the paddle and leaves. These results highlight the efficacy of the gating network. However, the gating network behavior is dependent on the dynamics of the game, which is evident in Table \ref{tab:gaze-usage}. For these games, we can see that the maximum utilization of gaze is under 40\% of the total number of frames seen during gameplay. 

\begin{table}[h]
    \centering
    \caption{Percentage of gaze usage over the entire duration in different games.}
    \label{tab:gaze-usage}
    \begin{tabular}{lcccccccc}
    \toprule
    \textbf{Game} &
      \textbf{Asterix} &
      \textbf{Breakout} &
      \textbf{Centipede} &
      \textbf{MsPacman}&
      \textbf{Phoenix} &
      \textbf{SeaQuest}
       \\ \midrule
    Gaze usage (\%) &
      0.368 &
      0.365 &
      0.369 &
      0.266 &
      0.227 &
      0.068 \\ \bottomrule
    \end{tabular}
\end{table}

\subsection{Action Prediction Network Results}
For action prediction, we train a network from the input image stack and predicted gaze on the full Atari action set of 18 actions, as described in section \ref{sec:act-pred-net}. We have used an independently trained network for gaze prediction (see \ref{subsec:gaze-prediction-network}). The predicted gaze is modulated by the gating network (section \ref{sec:gating-network}) before sending it to the action prediction network. The hyperparameters for network training are the same as the gaze prediction network.
We compare the SEA network to two different approaches discussed in \cite{zhang2019atari} and \cite{saran2020efficiently}. Though \cite{saran2020efficiently} presents an approach to implicitly include gaze information into the network, we hypothesize that explicit modeling of when to use the gaze provides a better way to understand game dynamics and action learning. For bench-marking, we use baselines described in \cite{saran2020efficiently}: Behaviour Cloning (BC), Attention Guided Imitation Learning (AGIL), and Random gated SEA. The random gated version is exactly the SEA network expect the gating function is not learnt. Instead, the output is randomly chosen as 1 or 0 from an uniform distribution. The same random behaviour is used during the learning and inference phase. 

\begin{table}[b]
  \centering
  \caption{Action classification accuracy of SEA in comparison with Majority Action, BC, and AGIL.}
    \begin{tabular}{lcccc}
    \toprule
    \textbf{Game} & \textbf{Majority Action} & \textbf{BC} & \textbf{AtariHEAD-AGIL} & \textbf{SEA} \\
    \midrule
    Asterix & 0.365 & 0.68 & 0.532 & \textbf{0.621} \\
    Breakout & 0.8 & 0.79 & \textbf{0.816} & 0.595 \\
    Centipede & 0.581 & 0.37 & \textbf{0.628} & 0.57.4 \\
    MsPacman & 0.266 & 0.555 & 0.678 & \textbf{0.681} \\
    Phoenix & 0.291 & 0.33 & \textbf{0.658} & 0.545 \\
    SeaQuest & 0.208 & 0.47 & \textbf{0.505} & 0.37 \\
    \bottomrule
    \end{tabular}%
  \label{tab:accuracy}%
\end{table}%

Regarding action classification accuracy, Table \ref{tab:accuracy} provides SEA network performance in comparison to the benchmark approaches. The SEA network performs well only in two games (Asterix and MsPacman), while the AGIL outperforms us in other four games. One of the caveats of using gating units like GRU is that it introduces additional parameters into the module and increases training time and complexity. We believe this to be one of the reasons behind lower game scores seen from SEA compared to AGIL.

It should be noted that action classification accuracy is not correlated with the game score which is the ultimate goal of the learning structure. 
There are two main fundamental differences in the nature of the gameplay and action classification that can potentially cause this performance mismatch. First, the gameplay is dynamical, i.e., actions depend on the previous state, while such a dependency is absent in the classification problem. Hence, the data of the classification is different from the data coming from game play. 
Second, in the gameplay there are often more than one viable potential action for a given situation. This is even seen in the actions recorded from different human subjects. In the gameplay the selection of each of the valid alternative actions will results in a score, while in the classification problem it will be counted as a miss and hence reducing the classification accuracy.

To shed more light on this matter, lets consider the game break-out.
Using a single ball, the player must use the paddle below to guide the ball knocking down as many bricks as possible. When the ball is moving upward the action taken will have no effect on the score, but it contributes to the classification accuracy. 
For this specific example we have observed about 10\% higher classification accuracy (50\% vs. 40.5\%) at the moments that actions have direct effect on the game score (ball moving downward vs. going upward). For some of the games this can be further generalized to taking any actions vs. taking no action at all in response to a certain situation.
In other words, the classification performance can be calculated by grouping the human-demonstrations as action and no-action which removes the discrepancy caused by different players taking different actions for the same game state. 

To further clarify, Table \ref{tab:f1-score} provides F1-scores when classification is done between no action and action, here the exact subject's action is irrelevant (hence high F1 score). However, when we consider what exact action subjects did, the F1 score decreases. This further validated that even though subjects took different actions, the game score was not affected and hence a good classification accuracy need not reflect a good game score or vice-versa.  

\begin{table}[h]
\renewcommand{\arraystretch}{1.15}
  \centering
  \caption{F1 scores of classification with all the actions and dropping no-action/invalid actions}
    \begin{tabularx}{\textwidth}{lcccccc}
    \toprule
    \textbf{Game} & \textbf{Asterix} & \textbf{Breakout} & \textbf{Centipede} & \textbf{MsPacman} & \textbf{Phoenix} & \textbf{SeaQuest}\\
    \midrule
    \textbf{Action vs. no action} & 0.95 & 0.76 & 0.93 & 0.98 & 0.90 & 0.97\\
    \midrule
    \textbf{All the actions} & 0.62 & 0.65 & 0.57 & 0.70 & 0.48 & 0.37\\
    \bottomrule
    \end{tabularx}%
  \label{tab:f1-score}%
\end{table}%

\section{Game Performance Analysis}

\begin{table}[b]
 \renewcommand{\arraystretch}{1.25}
  \centering
  \caption{Game scores of SEA in comparison with BC, AGIL, and Random gated SEA.}
  \resizebox{\linewidth}{!}{%
    \begin{tabular}{lccccccc}
    \toprule
    ~ & \textbf{SEA} & \multicolumn{2}{c}{\textbf{BC}}  & \multicolumn{2}{c}{\textbf{AGIL}} & \multicolumn{2}{c}{\textbf{Random SEA}}\\
    \midrule
    \textbf{Game} & Score & Score & p-value  & Score & p-value& Score & p-value\\
    
    \midrule
    Asterix & \textbf{608.3 \(\pm\) 148.3} &  246.7 \(\pm\) 166.8 & < 0.01 & 410 \(\pm\) 153.0& < 0.01 &  408.3 \(\pm\) 122.5 & < 0.01 \\
    
    Breakout & 7.13 \(\pm\) 2.68 & 1.76 \(\pm\)1.54& < 0.01 &  \textbf{7.26 \(\pm\) 1.61} & 0.77 & 3.4 \(\pm\) 1.36 &< 0.01\\
    
    Centipede & \textbf{13023 \(\pm\) 4333} & 528 \(\pm\) 331 &< 0.01& 12099 \(\pm\) 4512 &< 0.01&  7086 \(\pm\) 2376&< 0.01  \\
    
    MsPacman & \textbf{1258 \(\pm\) 385.7} & 265  \(\pm\)  85.5 &< 0.01& 1008 \(\pm\) 255.0  &< 0.01&  542 \(\pm\) 152.1&< 0.01 \\
    
    Phoenix & \textbf{4905 \(\pm\) 1106} & 4461 \(\pm\) 1361 &0.18  & 3503 \(\pm\) 1098 &< 0.01&  4019 \(\pm\) 751.8&< 0.01\\
    
    SeaQuest & \textbf{304.6 \(\pm\)43.7} & 104.00\(\pm\)  24.44 &< 0.01& 232.67 \(\pm\) 37.77 &< 0.01&  230.67 \(\pm\) 34.92&< 0.01  \\
    \bottomrule
    \end{tabular}%
    }
  \label{tab:game-scores}%
\end{table}%

The trained SEA network and all the baselines are evaluated thirty times for each game. During evaluation, the random seed is kept the same across SEA and other baselines. The averaged games score and the standard deviations over the thirty evaluations are listed in Table \ref{tab:game-scores}. Also, a Welch's t-test is conducted between the SEA game score with each of the other methods to analyze the statistical significance in that specific game and the p-values are listed in respective columns. It can be seen that SEA approach outperform all other approached. It should be noted that AGIL provides a slightly better average score in Breakout comparing to SEA but there is no statistical differences between the score of the two approached and therefore both AGIL and SEA have the highest score in Breakout.
The detailed comparison of SEA outcome to other baseline methods is discussed next.

\textbf{BC vs. SEA}:
In Behavior cloning (BC), the policy is a simple action imitation through a straight forward classification task of actions. SEA augments the action classification from game frames similar to BC with selective gaze information. We can emulate the BC approach in our model by keeping the gate-out as zero. The performance gain in SAE, when compared to BC, is because of the supplemented gaze information that helps to guide the system towards important aspects of the frame similar to \cite{zhang2019atari}. Interestingly, even randomly using the gaze information (Random gated SEA) approach outperforms the BC approach (Table \ref{tab:game-scores}) indicating the advantage of using the human-gaze to enhance the imitation learning. 

\textbf{AGIL vs. SEA}: Attention Guided Imitation Learning \cite{zhang2019atari} introduces explicit calculation of gaze and overlay of the convolved feature maps of the game frame to augment the simple method of Behaviour Cloning. From Table \ref{tab:game-scores} it is pretty evident that AGIL outperforms BC. 
However, SEA is developed on the main hypothesis that learning when to augment a game frame with the gaze information should outperform the augmentation at all time specially if gaze is directly integrated without considering its dynamic and type (fixation vs. saccade).
Consequently, we can see that SEA outperforms AGIL (Table \ref{tab:game-scores}). If the SEA model chooses to ignore all the gaze information, the model performance should fall back to Behaviour cloning (BC), and using all the gaze data should result in at least the performance of AGIL. It is even not surprising, to see the AGIL performance to be slightly higher that the random gated SEA network (even lower in the case of Phoenix game). This further question the overall benefits of using gaze augmentation blindly. 

\textbf{SEA vs Random gated SEA}:
Comparing the SEA to AGIL we showed the benefits of selective use of gaze for augmenting the learning process. However, it is also critical to show that our proposed gating network is indeed learning the instances to use the gaze information. This is done through the comparison of SEA with a random gated SEA. As seen in the Table \ref{tab:game-scores}, learning the instances to use the gaze outperforms random use of the gaze which also indicate that the learnt gating functions output is not random and depends on the game dynamics. 

\section{Conclusions}
In this work, we propose a selective eye-gaze augmentation of imitation learning in which the network learn when to use gaze information in enhancing the learning. 
It can be thought of as a more generalized version of simple Behaviour cloning (no eye-data) and Attention Guided Imitation Learning (AGIL, mask all the game frames with eye-data). By modulating the gating behavior, we can emulate both BC as well as AGIL networks. 

Comparing to the Behavioral cloning, even a random selection of gaze for enhancement of imitation learning resulted in a better performance. In comparison to Attention Guided Imitation Learning (AGIL) which masks all the game frames with eye-data, our method consistently performed better in several Atari games. 
The outcome of our evaluation studies not only indicates the benefits of gaze in enhancing the learning but also highlights the importance of selective gaze usage in the learning process. 
It should be noted that our gating network is independent of the task and its associated visual complexity. As a future direction, more evaluation studies may be conducted in the presence of games with higher level of visual complexity to further examine the performance of the proposed network. 

Finally, the present SEA architecture does not consider any temporal dependence. Hence, as future work, we plan to implement the SEA network (sub-networks) with temporal dependence using recurrent neural networks. We can also extend the SEA in to a reinforcement learning setting where the selective use of game frames can be learnt online.  

\begin{ack}
We gratefully acknowledge the support of NVIDIA Corporation with the donation of the Titan Xp GPU used for this research.
\end{ack}

\printbibliography

\end{document}